# Trade-off Between Spatial and Angular Resolution in Facial Recognition

Muhammad Zeshan Alam, Sousso kelowani, and Mohamed Elsaeidy

[1] Brandon University, Brandon MB, Canada
[2] University of Quebec at Trois Rivieres , 3351 Bd des Forges, Trois-Rivieres, Canada
[3] Istanbul Medipol University, Istanbul, Turkey
zeshusher@gmail.com

**Abstract.** Ensuring robustness in face recognition systems across various challenging conditions is crucial for their versatility. State-of-the-art methods often incorporate additional information, such as depth, thermal, or angular data, to enhance performance. However, light field-based face recognition approaches that leverage angular information face computational limitations. This paper investigates the fundamental trade-off between spatio-angular resolution in light field representation to achieve improved face recognition performance. By utilizing macro-pixels with varying angular resolutions while maintaining the overall image size, we aim to quantify the impact of angular information at the expense of spatial resolution, while considering computational constraints. Our experimental results demonstrate a notable performance improvement in face recognition systems by increasing the angular resolution, up to a certain extent, at the cost of spatial resolution.

**Keywords:** Face recognition, light field imaging, spatio-angular resolution, deep learning, Convolution neural networks

## 1 Introduction

Face recognition system are being rapidly adopted in wide range of applications including, surveillance, entertainment, and forensic etc. specially because it is a widely acceptable universal biometric modality. However, face recognition still encounter various challenges in a completely unconstrained environment involving significant variations in illumination, pose, emotions, and occlusions.

Recently, deep learning models have contributed significantly in the success of computer vision in general and face recognition in particular. Despite the tremendous potential of the deep learning models the aforementioned challenging scenario limits their performance. This is because learning based model obtain relevant features through training process in which the model weights are updated as the algorithm is fed the labeled data. The training data is typically a regular camera images which is heavily effected by these challenging environmental conditions and lose some of the spatial information critical in face discrimination.

Light field (LF)imaging, allows us to capture the additional angular information of the light which is lost in conventional imaging. This is usually achieved by separately recording the intensities of light rays coming from different directions are at each separate pixel position [1–3]. This additional angular information turns out to be a critical discriminatory factor for various classification tasks. For example Light fields used for face recognition have shown significant performance improvement [4, 5], particularly when combined with deep learning-based light field recognition techniques.

The performance improvement of the LF based face recognition methods however, involve additional computational cost either as a pre-processing step ( posteriori refocusing or depth computation) or due to some modification of network architecture in case of learning based methods [5], depending on the LF representation used in the method. The overhead involved in the processing cost of the additional angular information is one of the major limitation of the LF based techniques, however, acquiring the angular information also presents its own set of problems.

Light field imaging systems can be implemented in a variety of ways including, camera arrays [6, 7], coded mask LF camera designs, [8, 9], and microlens array (MLA) [10, 2] based cameras. In the camera array based approach, each camera separately captures the scene from a different perspective, at a high spatial resolution, however, camera arrays are expensive and have restricted mobility. An alternative approach for the light field acquisition is the coded mask-based camera design. There are several design approaches, including, single shot, single mask, multiple shot, and multiple mask-based camera designs, irrespective of the design coded masks methods have poor light efficiency. MLA-based LF cameras are the most cost-effective and easy to handle LF acquisition system. However, since a single sensor is used to capture both spatial and angular information, a spatio-angular resolution trade-off exit in all MLA based designs [11, 12].

To minimize the additional complexity associated with the angular information, we study the effect of this spatio-angular resolution trade-off on the performance of the learning based face recognition models. For this purpose, we increase the angular resolution at the cost of the spatial resolution while maintaining the overall size of the input image. We introduce the super pixels of various sizes (angular resolution), formed by combining pixels from multiple perspective images and shown in Figure 3, for training the state of the art CNN based face recognition network and evaluate their performance for different angular resolution input images. The major contributions of this paper are summarized as follows:

1. Quantitative evaluation of the spatio-angular resolution trade-off on the performance of face recognition
2. Application of a compact light field representation in face recognition for improved prediction accuracy.

The paper is organized as follows. In Section 2, we present related work on face recognition, particularly the light field-based CNN approaches in face recognition. We explain the application of the light field to the face recognition problem in Section 3. Light field representations for training deep learning models are presented in section 4. In section 5 we detail the training process and hyper-parameters. In sections 6 and 7 we present the experimental result and conclude the paper respectively.

## 2   Related Work

Traditional face recognition methods primarily focus on analyzing 2D images and extracting discriminative facial features for identification. Eigenfaces [13] introduced the concept of Principal Component Analysis (PCA) for face representation. Fisherfaces [13] extended this approach with Linear Discriminant Analysis (LDA) to improve discriminability. Local Binary Patterns (LBP) [14] provided an effective texture-based feature descriptor for robust face recognition.

Deep learning-based face recognition techniques, like other computer vision tasks, are becoming state-of-the-art. Numerous studies have focused on developing deep neural network architectures and training methodologies to improve recognition performance. In FaceNet [15] a siamese network is trained with triplet loss to learn highly discriminative face embeddings. VGGFace [16] presented a deep convolutional neural network (CNN) trained on a large-scale face dataset, demonstrating impressive recognition accuracy. ArcFace [17] introduced a novel loss function based on angular margin to enhance discriminative feature learning. This method explicitly optimizes the angular margin between different classes, making the learned embeddings more separable. In CosFace [18] margin-based loss function is proposed hat introduces an additive cosine margin to the standard softmax loss. By directly maximizing inter-class variations and minimizing intra-class variations, CosFace achieved improved discrimination of face embeddings. The approach demonstrated robustness against variations in lighting, pose, and expressions. Circle Loss by Sun et al. [19] introduced a new loss function based on angular similarity. By utilizing pairwise angular similarity, the Circle Loss optimizes the feature space to form compact and well-separated clusters. This approach achieved competitive performance and demonstrated the capability to handle large-scale face recognition tasks.

Cross-domain face recognition, dealing with recognizing faces across different domains or modalities, has gained attention in recent years [20] and [21]. [20] Addresses the problem of pose-invariant face recognition through a two-stream deep neural network architecture. The method incorporates both spatial and temporal information to capture pose variations in face images. By jointly modeling appearance and motion cues, the approach achieved robust recognition performance across different pose angles. In [21] a generative adversarial network (GAN)-based approach for thermal-to-visible face recognition is proposed. The method utilizes a conditional GAN framework to generate synthetic visible face images from thermal images. By learning the mapping between thermal and visible domains, the approach improved recognition performance in cross-modal face recognition scenarios.

Zhang et al. [4] proposed a light field-based approach for face recognition combining depth and intensity information. The integration of intensity and depth cues allowed for improved robustness against variations in pose, illumination, and occlusion, leading to enhanced face recognition performance. In [5] a multi-modal light field representation that combines depth and color cues for robust face recognition is introduced. This incorporation of depth and color cues was aimed to enhance the system's robustness to challenging conditions. In [22], a novel two-long short-term memory( LSTM) cell architecture that leverages spatial and angular information from a light field image is proposed. The new two-cell network is able to jointly learn from multiple sequences of light field perspective images simultaneously, targeting to create richer and more effective models for the face recognition task. A double-deep spatio-angular learning framework for light field-based face recognition, which is able to model both the intra-view/spatial and inter-view/view/angular information using two deep networks in the sequence is proposed in [23]. The proposed framework includes a LSTM recurrent network, whose inputs are VGG-Face descriptions, computed using a VGG-16 convolutional neural network (CNN)

## 3    Light field for improved Face recognition

Light field imaging allows for the capture of angular information from light rays originating from a scene point. Conventional cameras, as depicted in Figure 1(a), record incident light rays directly at the sensor, resulting in the loss of directional information. In contrast, light field cameras, such as the MLA-based cameras [2] and [10], shown in Figure 1(b), incorporate a micro-lens array (MLA) adjacent to the sensor. This arrangement enables the separate recording of light ray intensities passing through different points (sub-apertures) of the main lens and converging at the micro-lens in front of specific pixels. Consequently, light field cameras preserve the directional information of the captured light rays.

The MLA in light field cameras, as illustrated in Figure 1(b), receives light rays reflected from scene points at varying depths with different incident angles [24]. This angular difference leads to disparities in the multiple perspective images obtained from decoding the light field. By tracing the light rays back to the scene, it becomes possible to calculate the corresponding depths of these points. Estimating the depth of all captured points generates a depth map that reveals the depth variations among different objects in the scene. Exploiting this depth information is valuable for facial analysis, as facial features like the nose, ears, and cheekbones exhibit substantial depth variations across individuals. Leveraging this depth information can assist face recognition in challenging conditions, such as pose changes and occlusions, enhancing its performance.

## 4    Spatio-angular resolution trade-off

Light fields can be represented in many ways, including, sequence of perspective images, lenslet representation, EPIs, and plenoptic function. Irrespective of the LF representation, incorporating the additional angular information results in the fundamental spatio-angular resolution trade-off. Typically a high spatial-resolution face image contains fine details, such as the texture of the skin, shape of facial features, and subtle patterns. These details are crucial for accurately distinguishing and identifying individuals. They provide more information for the face recognition algorithm to analyze and match against reference images.

Face recognition algorithms usually extract various facial features, such as the distance between the eyes, shape of the nose, or curvature of the lips, to create a unique face template or facial signature. With higher spatial-resolution images, these features can be extracted more accurately and reliably, leading to better recognition performance. While the high spatial resolution is beneficial, other factors, such as lighting conditions, pose, and partial occlusion, also play crucial roles in accurate face recognition. These afore-mentioned challenging factors can affect the fine details captured in the high-resolution images resulting in the loss of sufficient discriminative features. However, these spatial features, when complemented with features extracted from angular information such as facial features depth variation and Iris reflection etc, could provide robust face recognition cues in challenging conditions. Current State-of-the-art face recognition methods adopt deep CNN models that process images pixel-wise, meaning that each pixel is processed individually or in small receptive fields. Including both spatial and angular information in an input image means more pixels in the input image. With an increase in the number of pixels, the number of computations required to process the image also increases. More importantly, CNNs heavily rely on convolutional operations, that obtain relevant features through training using layers that consist of multiple kernels. These kernels are updated as the algorithm is fed labeled data, converging by numerical optimization methods on the weights that best match the training data[25]. With a larger input image size, the convolutional operations performed in each layer increases, which results in an increase in the number of multiplications and additions required for the convolutional operations, leading to higher computational complexity. The size of the output feature map after the convolution operation in each layer depends on three factors: the size of the input image, the size of the kernel, and the stride. The stride specifies the step size at which the kernel moves across the input image.

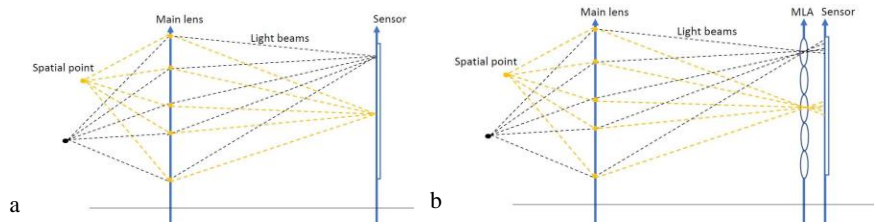

Fig. 1: An illustration of MLA-based light field camera image acquisition in comparison with conventional camera model. (a) Conventional camera model. (b) MLA-based light field camera capturing scene points at different depths.

Light fields which are typically represented as a set of up to one hundred perspective images with standard spatial resolution would render any real-time surveillance system computationally prohibitive and could also exhaust the memory in embedded systems. Other light field representations discussed below would also suffer from computational cost issues due to separate pixels recording of spatial and angular information:

### 4.1 perspective or sub-aperture image representation

A sub-aperture representation refers to a subset of images extracted from the captured light field data. In addition to the spatial information of the scene, Sub-aperture images contain angular information of the light as they are captured using an array of micro-lenses or camera units that are positioned at different viewpoints. Each micro-lens or camera unit captures a subset of the rays coming from various directions within the scene, which corresponds to different angular perspectives. To capture the light field, the micro-lenses or camera units are arranged in a grid pattern, covering the sensor or image plane. Each micro-lens or camera unit captures a small portion of the incoming light rays, corresponding to a specific angular perspective. By combining the captured data from all the micro-lenses or camera units, a complete light field is obtained. In the resulting sub-aperture images, each image represents a specific viewpoint or perspective within the light field. The angular information is encoded in the differences between these viewpoints. By analyzing the parallax or disparity between the sub-aperture images, it is possible to extract depth information and estimate the relative distances of various facial features, which serve a robust cue in face identification under challenging conditions.

## 4.2   Epipolar Plane Image representation

In the Epipolar Plane Image (EPI) representation, the light field data is arranged in a specific format that emphasizes the epipolar geometry. Epipolar lines are the lines of sight connecting corresponding points in different viewpoints. By plotting these lines and their associated image intensities, an EPI is formed. To create an EPI, one dimension of the EPI corresponds to the spatial information, typically representing the pixel rows or columns, while the other dimension corresponds to the angular information, representing the viewpoints or perspectives. Each point in the EPI corresponds to the intensity of a specific pixel in the captured sub-aperture images. By examining the EPI, it is possible to observe the disparities or shifts in intensity values along the epipolar lines, which relate to the depth or parallax information in the scene. However, it should be noted that EPI is an alternative arrangement of the pixels that allows better visualization of the depth but does not address the spatial-angular trade-off and therefore suffer from high computational complexity issues.

## 4.3   Lenslet representation

The raw lenslet light field representation refers to the raw sensor data captured by the individual micro-lenses. The raw lenslet light field representation consists of a grid of sub-images or patches, where each sub-image corresponds to the image captured by a specific micro-lens. A lenslet is an individual component that captures a subset of light rays, while a sub-aperture image is whole resulting image obtained from the captured light field data. The lenslet image actually represents a magnified view of the captured scene.

To create the lenslet image representation of the light field, the lenslet images are arranged in a grid pattern, similar to the arrangement of the micro-lenses. The lenslet images are combined to form a larger mosaic that represents the entire light field. However, like a set of perspective images or EPI representation, lenslet image representation also suffers from the spatio-angular resolution trade-off.

### 4.4 Proposed Representation

In this section, we introduce a lenslet-inspired representation of the light field to explore the significance of both spatial and angular information in the context of face recognition under difficult conditions. To achieve this, we modify the middle perspective image by replacing individual pixels with macro-pixels that contain multiple perspectives of the same scene point. To maintain the overall size of the original input image we drop the same number of neighboring spatial pixels of the pixel that is replaced by the macro-pixel. Considering that the decoded image has dimensions of 625 x 434 pixels, we construct macro-pixels as multiples of 2 pixels, as illustrated in Figure 3. The selection of neighboring perspective pixels included in each macro-pixel depends on their distance from the middle perspective. For instance, a 2x2 macro-pixel encompasses the four adjacent perspective image pixels. By increasing the size of the macro-pixels, we prioritize the contribution of angular information in the light field representation, albeit at the expense of spatial information. This enables us to investigate the significance of angular cues in comparison to spatial features for distinguishing individual faces, particularly under challenging conditions.

## 5 Training

We trained two different CNNs namely GoogleNet and Resnet50 and modified the final layers to match the fifty class output of the [22] dataset. For GoogleNet, we removed the 'loss3-classifier', 'prob', 'output' layers and added a fully connected(FC) layer for 50 classes, a softmax, and a classification layer. For Resnet50 we replace the fully connected, 'softmax', and 'classification' layers with the corresponding layer for 50 classes.

Both CNNs are pre-trained on ImageNet dataset [26] and transfer learning is performed to adapt and fine-tune pre-trained Googlenet and Resnet50 for the face recognition problem. The IST-EURECOM Light Field Face Database (LFFD) [22] consists of LF face images captured by a Lytro ILLUM camera and is used here for performance assessment purposes. The IST-EURECOM LFFD includes 4000 LF images, captured from 50 subjects, in two separate acquisition sessions with a temporal separation between 1 and 6 months. Each session contains 20 LF images per subject with different facial variations including facial expressions, poses, illuminations, and occlusions, as illustrated in Figure 2. We have separated on type of facial variations LF images of all 50 subjects for testing and the rest of the LF images constitute the training dataset.

The networks performances are robust under different training hyper-parameters configurations. Various combinations of the learning rate, optimizer, and batch size, etc are tested but the variations in performance are negligible. The batch size turned out to be the most influential hyper-parameter in terms of performance improvement for both the networks described above. Increasing the batch size improved the performance and the maximum batch size supported due to memory limitations is 35.

Among different configurations, the following set of hyperparameters resulted in the best performance and therefore, adopted in this work. The initial learning rate is set to 3e-4 with a learning rate drop of 0.1 and a learning rate drop period of 20. Although the choice of optimizer has shown a marginal effect on the overall performance, However, adam optimizer has slightly improved the prediction and hence selected in our training. The batch size is set to 32 and the training is performed for 50 and 100 epochs for Resnet50 and Googlenet respectively.

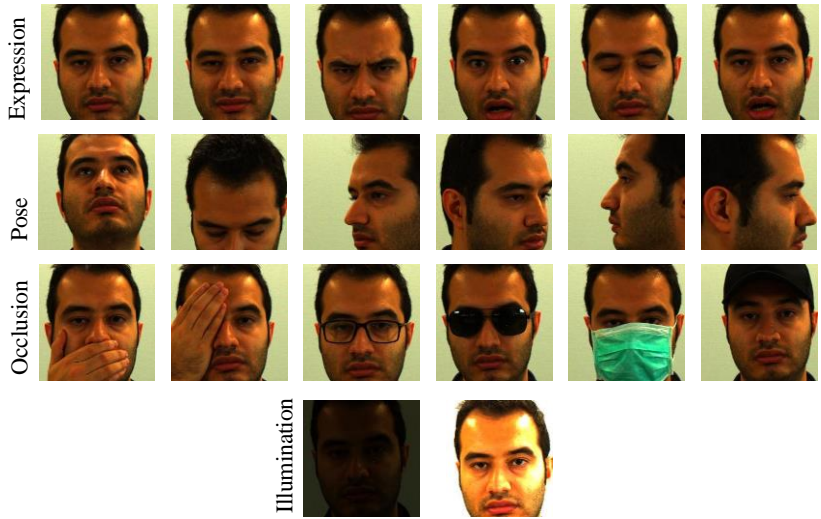

Fig. 2: Visualization of the four different categories of facial variations in the data-set.

## 6    Experimental Results

In this section, we present the results of the spatial-angular resolution trade-off in face recognition. We provide the accuracy of prediction by categorizing the various facial variation into four distinct groups which typically pose challenges in face recognition.

In Figure 4, a quantitative comparison of the proposed macro-pixel LF representation at different angular resolutions using Resnet50 architecture is presented. It can be seen in Figure 4 that macro-pixel representation outperforms regular 2D image-based face recognition for almost all types of facial variations. However, there is no uniform increase in the performance based on the increase in the angular resolution. On the other hand with the significant increase in the macro-pixel size, the impact of decrease in the spatial resolution on the overall performance tends to dominate and causes a drop in the prediction accuracy. Additionally, for more challenging scenarios such as pose variation, illumination, and occlusion the benefit of angular resolution is more evident than the regular two-dimensional images and for more commonly encountered scenarios like change in facial expression, there is no additional benefit of angular resolution after the network convergence. Several deep CNN architectures designed for image classification exist in the literature. The performance of these CNN models is influenced by many design choices, for example, the number of trainable parameters, Depth, and width of architecture [27]. To demonstrate the robustness of the angular information in lieu of spatial information the results for a different CNN architecture Googlenet are presented in Figure 5.

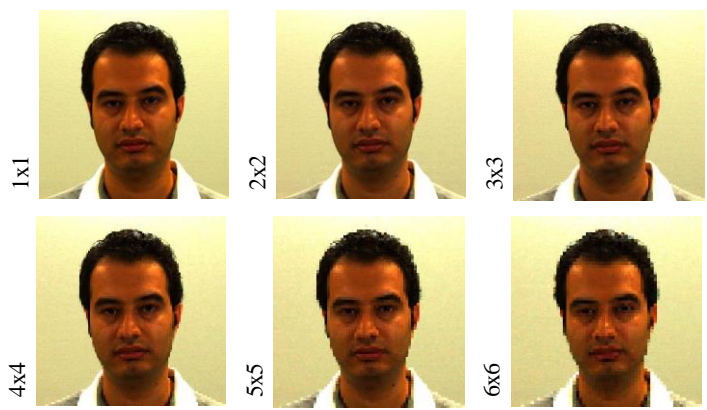

Fig. 3: Proposed image representation for Spatio-angular resolution trade-off investigation. Angular resolution varies from 1x1 to 6x6.

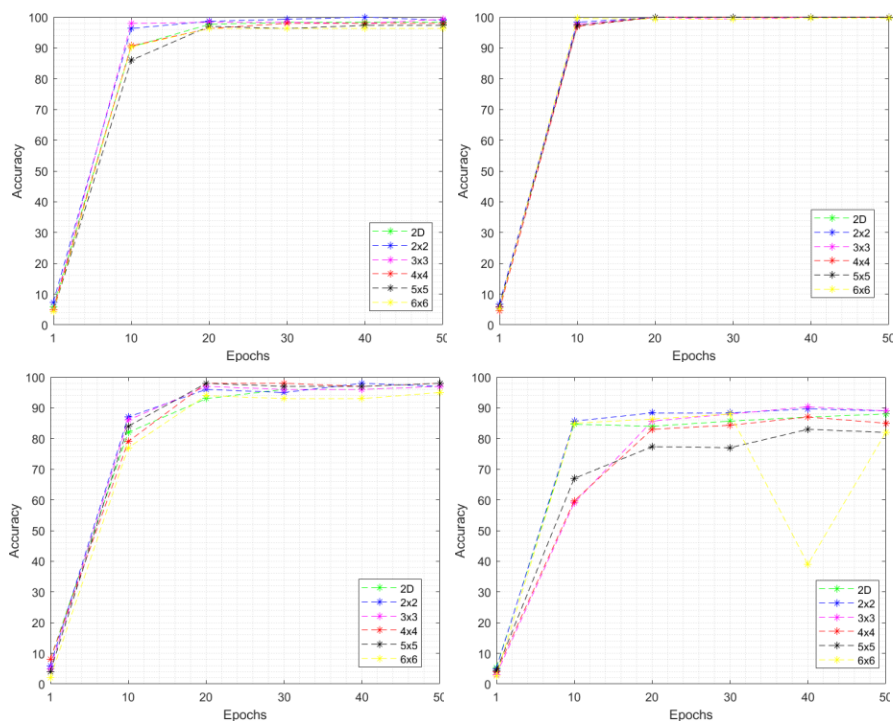

Fig. 4: Performance comparison of the spatial-angular resolution trade-off with different angular resolutions macro-pixels. Top Left) Occlusion. Top Right) Expression. Bottom Left) Illumination. Bottom Right) Pose.

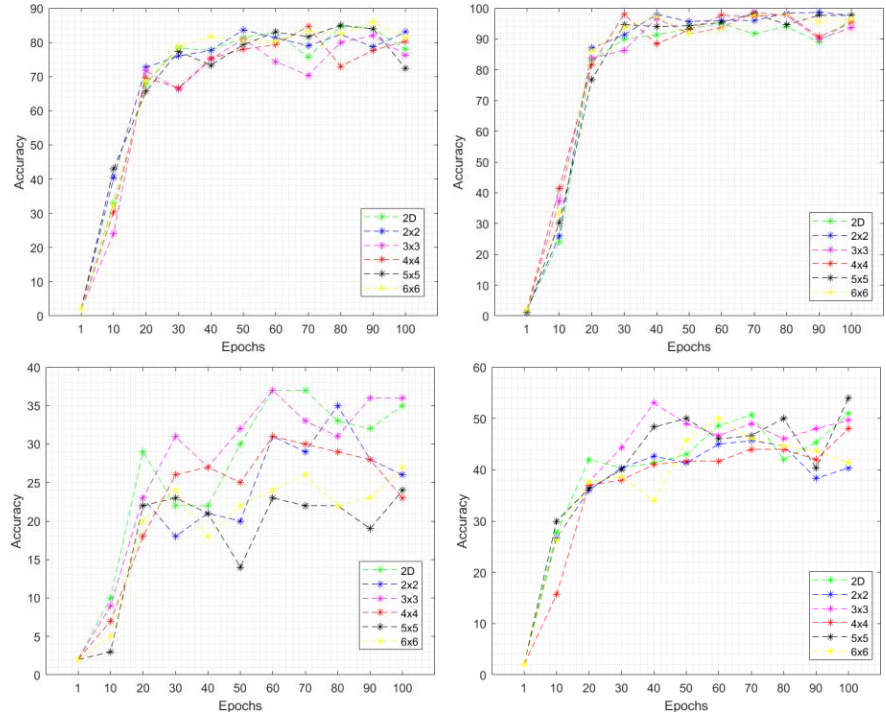

Fig. 5: Performance comparison of the spatial-angular resolution trade-off with different angular resolutions macro-pixels. Top Left) Occlusion.Top Right) Expression. Bottom Left) Illumination. Bottom Right) Pose.

Resnet50 consists of 50 layers, designed to tackle the challenge of training very deep networks effectively. It introduces a concept called residual connections, where shortcut connections are added to skip over certain layers, allowing the model to learn residual functions.The ResNet50 architecture has approximately 23.5 million trainable parameters. GoogleNet, on the other hand, is 22 layers deep Architecture, with 27 pooling layers included. The overall architecture includes 9 linearly stacked inception modules, with a total of seven million trainable parameters.

Although the two networks are widely different in their design approach, in Figure 4, it can be seen that the angular information proves to be robust in the aforementioned challenging scenarios despite the loss of spatial information. However, in Figure 5, an increase in the macro-size pixel seems also to benefit the more common scenario of regular facial expression variation but has no apparent effect on the illumination variation condition. Additionally, Resnet50 tends to be sensitive to the loss of spatial information and therefore, encounters a loss in performance for large macro-pixel sizes of 5x5 and 6x6 but google net, shows an overall gain in the performance for high angular resolution.

## 7    Conclusion

This paper delves into the trade-off between spatial and angular resolution in face recognition, focusing on the lenslet-inspired macro-pixel representation. The proposed LF representation showcased performance improvements when compared to conventional image-based methods. However, it becomes apparent that there exists a threshold for increasing angular resolution while sacrificing spatial resolution. Furthermore, it is important to highlight that even with notable advancements in face recognition under normal conditions, the proposed LF representation can offer advantages in challenging scenarios without increasing computational costs.

## References

1. M. Levoy and P. Hanrahan. Light field rendering. In *ACM Int. Conf. on Computer Graphics and Interactive Techniques*, pages 31–42, 1996.
2. R. Ng, M. Levoy, M. Brédif, G. Duval, M. Horowitz, and P. Hanrahan. Light field photography with a hand-held plenoptic camera. In *Stanford University Computer Science Technical Report CSTR*, 2005.
3. M. Z. Alam and B. K. Gunturk. Hybrid stereo imaging including a light field and a regular camera. In *Signal Processing and Communication Application Conference*, pages 1293–1296. IEEE, 2016.
4. P. Sepas, A.and Correia, K. Nasrollahi, T. B. Moeslund, and F. Pereira. Light field based face recognition via a fused deep representation. In *2018 IEEE 28th International Workshop on Machine Learning for Signal Processing (MLSP)*, pages 1–6, 2018.
5. Zhang X. Li S. Wang, Z. Multi-modal light field representation for robust face recognition. In *IEEE Access*, pages 1280–1288, 2019.
6. B. Wilburn, N. Joshi, V. Vaish, E. V. Talvala, E. Antunez, A. Barth, A. Adams, M. Horowitz, and M. Levoy. High performance imaging using large camera arrays. *ACM Trans. on Graphics*, 24:765–776, 2005.
7. J. C. Yang, M. Everett, C. Buehler, and L. McMillan. A real-time distributed light field camera. In *Eurographics Workshop on Rendering*, pages 77–86, 2002.
8. M. Z. Alam and B. K. Gunturk. Deconvolution based light field extraction from a single image capture. In *IEEE Intl. Conf. on Image Processing*, pages 420–424, 2018.
9. A. Ashok and M. A Neifeld. Compressive light field imaging. In *SPIE Defense, Security, and Sensing*, volume 7690, 2010.
10. M. Z. Alam and B. K. Gunturk. Hybrid light field imaging for improved spatial resolution and depth range. *Machine Vision and Applications*, 29:11–22, 2018.
11. M. Z. Alam and B. K. Gunturk. Light field extraction from a conventional camera. *Signal Processing: Image Communication*, 109:116845, 2022.
12. Javeria Shabbir, M Zeshan Alam, and M Umair Mukati. Learning texture transformer network for light field super-resolution. *arXiv preprint arXiv:2210.09293*, 2022.
13. M. Turk and A. Pentland. Eigenfaces for recognition. *Journal of cognitive neuroscience*, 3(1):71–86, 1991.

14. T. Ahonen, A. Hadid, and M. Pietikainen. Face description with local binary patterns: Application to face recognition. *IEEE transactions on pattern analysis and machine intelligence*, 28(12):2037–2041, 2006.

15. F. Schroff, D. Kalenichenko, and J. Philbin. Facenet: A unified embedding for face recognition and clustering. In *Proceedings of the IEEE conference on computer vision and pattern recognition*, pages 815–823, 2015.

16. O. M. Parkhi, A. Vedaldi, and A. Zisserman. Deep face recognition. 2015.

17. J. Deng, J. Guo, N. Xue, and S. Zafeiriou. Arcface: Additive angular margin loss for deep face recognition. In *Proceedings of the IEEE/CVF conference on computer vision and pattern recognition*, pages 4690–4699, 2019.

18. H. Wang, Y. Wang, Z. Zhou, X. Ji, D. Gong, J. Zhou, Z. Li, and W. Liu. Cosface: Large margin cosine loss for deep face recognition. In *Proceedings of the IEEE conference on computer vision and pattern recognition*, pages 5265–5274, 2018.

19. Y. Sun, C. Cheng, Y Zhang, C. Zhang, L. Zheng, Z. Wang, and Y. Wei. Circle loss: A unified perspective of pair similarity optimization. In *Proceedings of the IEEE/CVF conference on computer vision and pattern recognition*, pages 6398–6407, 2020.

20. L. Zheng, S. Wang, X. Liu, and Q. Tian. Pose-invariant face recognition with multi-view deep representation. In *IEEE Transactions on Pattern Analysis and Machine Intelligence*, pages 2290–2302., 2020.

21. T. Zhang, A. Wiliem, S. Yang, and B. Lovell. Tv-gan: Generative adversarial network based thermal to visible face recognition. In *2018 international conference on biometrics (ICB)*, pages 174–181. IEEE, 2018.

22. A. Sepas-Moghaddam, Al. Etemad, F. Pereira, and P. Correia. Long short-term memory with gate and state level fusion for light field-based face recognition. *IEEE Transactions on Information Forensics and Security*, 16:1365–1379, 2021.

23. Alireza S., M. A. Haque, P. L. Correia, K. Nasrollahi, T. B. Moeslund, and F. Pereira. A double-deep spatio-angular learning framework for light field-based face recognition. *IEEE Transactions on Circuits and Systems for Video Technology*, 30(12):4496–4512, 2020.

24. A. Wahab, M. Z. Alam, and B. K. Gunturk. High dynamic range imaging using a plenoptic camera. In *Signal Processing and Communications Applications Conference*, pages 1–4, 2017.

25. M. Z. Alam, S. Kelouwani, J. Boisclair, and A. Amamou. Learning light fields for improved lane detection. *IEEE Access*, 11:271–283, 2023.

26. A. Krizhevsky and G. Hinton. Imagenet classification with deep convolutional neural networks. *Advances in neural information processing systems*, pages 1097–1105, 2012.

27. M. Z. Alam, H. F Ates, T. Baykas, and B. K. Gunturk. Analysis of deep learning based path loss prediction from satellite images. In *Signal Processing and Communications Applications Conference*, pages 1–4. IEEE, 2021.